\documentclass[11pt,a4paper]{article}
\usepackage[hyperref]{ranlp2021}
\usepackage{times}
\usepackage{latexsym}

\usepackage{subcaption}
\usepackage{array}
\usepackage{amsmath}
\usepackage{adjustbox}
\usepackage{booktabs}
\usepackage{bbm}
\usepackage{babel}
\usepackage{graphicx}
\usepackage{amssymb}
\usepackage{epstopdf}
\usepackage{enumerate}
\usepackage[utf8]{inputenc}
\usepackage{caption}
\usepackage{amsmath}
\usepackage{multirow}
\aclfinalcopy 
\def\aclpaperid{97} 

\newcommand{\akshay}[1]{}
\renewcommand{\akshay}[1]{{\color{blue} AG: {#1}}}

\newcommand{\mahak}[1]{}
\renewcommand{\mahak}[1]{{\color{red} MG: {#1}}}

\newcommand{\specialcell}[2][c]{%
  \begin{tabular}[#1]{@{}c@{}}#2\end{tabular}}

\title{A Dynamic Head Importance Computation Mechanism for Neural Machine Translation}

\author{Akshay Goindani \qquad  Manish Shrivastava  \\International Institute of Information Technology - Hyderabad \\ \{\texttt{akshay.goindani@research.iiit.ac.in, m.shrivastava@iiit.ac.in}\}}

\begin{document}
\maketitle
\begin{abstract}

Multiple parallel attention mechanisms that use multiple attention heads facilitate greater performance of the Transformer model for various applications e.g., Neural Machine Translation (NMT), text classification.
In multi-head attention mechanism, different heads attend to different parts of the input. However, the limitation is that multiple heads might attend to the same part of the input, resulting in multiple heads being redundant. Thus, the model resources are under-utilized.
One approach to avoid this is to prune least important heads based on certain importance score. In this work, we focus on designing a Dynamic Head Importance Computation Mechanism (DHICM) to dynamically calculate the importance of a head with respect to the input. Our insight is to design an additional attention layer together with multi-head attention, and utilize the outputs of the multi-head attention along with the input, to compute the importance for each head. Additionally, we add an extra loss function to prevent the model from assigning same score to all heads, to identify more important heads and improvise performance. We analyzed performance of DHICM for NMT with different languages. Experiments on different datasets show that DHICM outperforms traditional Transformer-based approach by large margin, especially, when less training data is available.

\end{abstract}

\section{Introduction}
\label{introduction}
Transformer based NMT systems perform well on multiple translation tasks \citep{vaswani2017attention}. 
Multi-head attention is a very important component of the Transformer model~\citep{vaswani2017attention}. 
Multiple heads improve performance compared to a single head, as they allow the model to jointly look at different subspaces, and hence capture enhanced features from sentences.
For example, 
a head can capture positional information by attending to adjacent tokens, or it can capture syntactic information by attending to tokens in a particular syntactic dependency relation \citep{voita2019analyzing}.
%
However, the performance of the transformer-base model with 8 heads at each layer is only 1 BLEU point higher than that of a similar model with just a single head at each layer \citep{voita2019analyzing}. 
This is due to the fact that majority of the heads learn similar weights, and therefore, multiple heads attend to the same parts of the input. Hence, most of the heads are redundant, leading to an increased computational complexity without improving performance. 

To avoid this redundancy, one approach is to prune the redundant heads based on certain importance score. In this work, we focus on designing an importance computation method to compute the importance score for each head.
Some recent work has
analyzed the importance of heads 
by considering average attention weights of each head at some specific position \citep{voita2018context}. However, average of attention weights is a static measure of the head importance as it does not consider the varying importance of each head with respect to the input.
The importance of a head is dynamic, as a head can be very important for a particular word, but can be less important for other words. 
Thus, in this work, we propose a Dynamic Head Importance Computation Mechanism (DHICM) to calculate the importance score for each head, and this can be later utilized to design a pruning strategy.
Our key idea is to apply a second level attention on the outputs of all heads, to dynamically calculate the importance score for each head, that varies with the input, while training. 
{We also propose to add a new loss term to prevent our approach from assigning equal importance to all heads. Note that we apply DHICM for both self attention heads and encoder-decoder attention heads present in the encoder and decoder of the transformer architecture.}

To evaluate the performance of our method, we considered multiple translation tasks with different language pairs such as Hindi-English, Belarusian-English, German-English. 
Results show that DHICM achieves a much higher performance compared to the standard transformer model, particularly, in low-resource conditions where much less training data is available. 
Moreover, DHICM requires only ${\sim}d^2$ additional parameters ($d$ is the word embedding dimension), that is much less than the total number of parameters in the transformer base model. 
{The transformer model has a large number of hyperparameters, due to which, it is computationally challenging to search for their optimal values. 
Thus, much of the previous work used default values of the hyperparameters \citep{gu2018meta, aharoni2019massively}. However, these are not guaranteed to yield optimal performance on different datasets. Grid search over all hyperparameters is computationally intensive due to the exponential number of combinations across all possible values. 
Therefore, in this work, we perform grid search over a subset of hyperparameters, i.e., {\textit{architecture} hyperparameters} and {\textit{regularisation} hyperparameters}, 
and experiments show that the hyperparameter values obtained from our method yield significantly better performance compared to the default values.} 
{To summarize, our work makes the following major contributions:
\begin{itemize}
    \item We propose a Dynamic Head Importance Computation Mechanism for transformer based NMT systems, to compute the importance scores for all heads dynamically with respect to an input token.  
    \item 
    We propose to add an additional loss function that helps to compute different attention for different heads, and filter the most important heads.
    \item Our hyperparameter tuning method yields significantly better  performance than the default values. 
\end{itemize}
}

\section{Background}
\subsection{Single-Head Attention} \label{single_head_attention}

Given a sequence of $N$ $d$-dimensional vectors $X = (x_1,x_2,...,x_N)$ and a query vector $y \in \mathbb{R}^d$, a single-head attention is a weighted aggregate of $x_i$, $i \in \{1,2,...,N\}$, followed by a linear transformation. The weights are obtained using a function $F(x_i, q)$ e.g., multi-layer perceptron \citep{bahdanau2014neural} or scaled dot product \citep{vaswani2017attention}, 
and the attention $A_h(X, y|W_v,W_o)$ is computed as $A(X, y) = W_o \: {\sum\limits}_{i=1}^{N}F(x_i, y)W_vx_i$, where $W_o$ and $W_v$ are learnable weights. {In a transformer based NMT system, there is an encoder and a decoder. The encoder encodes the input sequence of tokens and outputs a sequence of vectors $X$. The decoder uses $X$ to generate a sequence of tokens. If the query vector $y$ is generated using the encoder, then the computed attention is known as self-attention. Whereas if the query vector $y$ is generated from the decoder, then the computed attention is known as encoder-decoder attention.}

\subsection{Multi-Head Attention} \label{multi_head_attention}

Multi-head attention mechanism runs through multiple single head attention mechanisms in parallel \citep{vaswani2017attention}.
Let there be a total of $H$ heads, where each head $h \in \{1,2,...,H\}$ corresponds to an independent single head attention. The output of each head $A_h(X, y|W_v^h,W_o^h)$ is calculated independently,
and the final output of multiple heads is calculated using the outputs of all heads, i.e., $\Sigma_{h=1}^{H}A_h(X, y|W_v^h,W_o^h)$,
where, $W_v^h,W_o^h$ are learnable weights for each head $h$. 

%

\section{Approach}

\subsection{Dynamic Head Importance Computation Mechanism (DHICM)} 
\label{DHIC}

In the traditional transformer model, the output of the multi-head attention is a linear transformation over the concatenation of outputs of all heads. 
Therefore, the outputs of all heads have equal contribution. However, 
since all heads are not equally important to the input (Sec.~\ref{introduction}), 
we propose to compute the importance of each head with respect to the input dynamically. 

Our idea is that an additional attention layer will allow the model to pay more attention to the head that is more important to the input. 
Thus, we design a second level attention 
that uses the input and output of all heads to compute attention scores, i.e., importance for all the heads 
with respect to the input, described as follows. 
Let $x \in \mathbb{R}^d$ be a $d$-dimensional input to the multi-head attention module, and $O^{h}$ be the output of head $h \in \{1,2,...,H\}$ (without applying the linear transformation $W_o^h$ described in Sections \ref{single_head_attention} and \ref{multi_head_attention}). 
We first learn a function $G(x, O^h)$ 
to determine the attention, i.e., importance score for head $h$.
To approximate $G(x, O^h)$, we considered both multi layer perceptron and scaled dot product.
In our experiments, we observed that both achieve similar performance, and since scaled dot product requires less number of parameters, we used the latter to compute $G(x, O^h)$:


\begin{equation}
    G(x, O^h) = \frac{\exp^{s(x, O^h)}}{\Sigma_{n=1}^{H}\exp^{s(x, O^n)}} 
\end{equation}
\noindent
where, 
\begin{equation}
     s(x, O^h) = \frac{{O^h}^{T}W^{T}Ux}{\sqrt d_{m}}
     \label{sx}
\end{equation}

\noindent
Here, $W \in \mathbb{R}^{d_m \times d_k}, U \in \mathbb{R}^{d_m \times d}$ are learnable parameters, and $d_k, d_m$ are scaling factors for the multi-head attention and second level attention, respectively. We also add a dropout layer \citep{srivastava2014dropout} after computing $Ux$ in Equation~\ref{sx}. Next, we compute the output of the second-level attention layer ($DHICM$) using the attention scores for each head, as follows: 
\begin{equation}
    DHICM(x, O) = W_s \: {\sum\limits}_{h=1}^{H}G(x, O^h)VO^h 
\end{equation}

\noindent
where, $O=(O^1,O^2,...,O^H)$, and $V \in \mathbb{R}^{d_m \times d_k}$ and $W_s \in \mathbb{R}^{d \times d_m}$ are learnable parameters. The output of the second level attention is then passed to the feed forward network. {Note that DHICM learns only ${\sim}d^2$ additional parameters corresponding to $W, U, W_s, V$ in the second layer added, and this is much less than the total number of parameters in the standard transformer model (typical value of $d$ is 512).} 

\paragraph{Objective} \label{Loss}
Let $L_c$ represent the cross entropy loss that is minimized to ensure that the model generates accurate tokens. However, by only considering $L_c$ as the objective, it might be possible that the model learns equal values of $G(x, O^h)$ for all $h \in \{1,2,...,H\}$. This would indicate that all heads are equally important to the input $x$, and thus, prevent us from filtering the most important heads. 
To avoid this, we add an extra loss term to penalize the model if the value of $G(x, O^h)$ becomes equal for all $h \in \{1,2,...,H\}$. 
More formally, let $a \in R^H$ be a vector representing the importance score of all heads according to the model, where $a_h = G(x, O^h)$ 
is the importance score of head $h$.
Let $b \in R^H$ be a vector representing equal importance of all heads, i.e., $b_h = \frac{1}{H}$, where $H$ is a constant. 
Both $a$ and $b$ denote the importance distribution of the heads, where $a$ is learned by the model using the second level attention, and $b$ is a uniform distribution with equal importance for all heads. 
In order to avoid the model from assigning equal importance to all the heads, we maximize the Kullback-Leibler divergence (KL Divergence) between distributions $a$ and $b$. Note that both the distributions sum up to 1, i.e., $\sum\limits_{h}a_h = 1$, and $\sum\limits_{h}b_h = 1$, and that $a_h > 0, b_h > 0$ for all $h \in \{1,2,...,H\}$.
Specifically, we add an extra loss term $L_{KL}$ as the KL Divergence between $a$ and $b$, given as:
\begin{equation}
    L_{KL} (a||b) = \sum\limits_{h \in \{1,2,...,H\}} a_h ln \frac{a_h}{b_h}
\end{equation}

\noindent The overall loss $L$, where we minimize $L_c$ and maximize $L_{KL}$, is computed as:

\begin{equation}
    L = L_c - \lambda * L_{KL}
\end{equation}

\noindent
where $\lambda$ is a hyperparameter used to control the effect of $L_{KL}$ on the overall loss $L$. The objective is to minimize the overall loss $L$. 

\section{Experiment}

\subsection{Dataset Description} \label{data}

\begin{table}[!t]
\centering
{
\begin{tabular}{ccccc}
	\toprule
	Dataset & Train & Validation & Test \\
	\midrule
	IWSLT14 & 160K & 7.3K & 6.7K \\
    WMT17-CS  & 5.9M & 3K & 6K \\
    HindEnCorp & 256K & 7K & 7K \\
    \specialcell{TED talks \\ (Be-En)} & 4.5K & 1K & 2.6K \\
	\bottomrule
\end{tabular}
}
\caption{Train, Validation and Test split size for different datasets used in our experiments}
\label{datasets}
\end{table}

\begin{table*}[!t]
\centering
{
\begin{tabular}{rrrrr}
	\toprule
	& \multirow{2}{*}[-0.5\dimexpr \aboverulesep + \belowrulesep + \cmidrulewidth]{{Default}} & \multicolumn{3}{c}{Optimal} \\
	\cmidrule(l){3-5}
	& & De-En & Hi-En & Be-En \\
	\midrule
	Feed forward dim. &  2048 & 2048 & 1024 & 128 \\
	Attention heads & 8 & 4 & 4 & 2 \\
	Dropout & 0.1 & 0.5 & 0.3 & 0.1\\
	Attention Dropout & 0.0 & 0.1 & 0.0 & 0.0 \\
	Activation Dropout & 0.0 & 0.3 & 0.0 & 0.0 \\
	Dropout (Section~\ref{DHIC}) & N/A & 0.5 & 0.2 & 0.2 \\
	Label Smoothing & 0.1 & 0.1 & 0.1 & 0.4 \\
	\bottomrule
\end{tabular}
}
\caption{Default and Optimal Hyperparameters}
\label{hyperparameters}
\end{table*}

We used German-English (De-En) parallel corpus obtained from IWSLT14 \citep{cettolo2014report} and WMT17 \citep{bojar2017results} shared translation tasks to evaluate the performance of our proposed method.
{Table~\ref{datasets} reports the number of parallel sentences in training, validation and test splits of different datasets that are considered in our experiments}.
To compare with \citep{iida2019attention}, 
we used WMT17 De-En training corpus as training set and newstest13 as validation set. Similar to \citep{iida2019attention}, we concatenated newstest14 and newstest17 to make one test set. We call this WMT17 dataset with the modified test set as WMT17-CS dataset.
To assess the performance of our method for low resource language pairs, we used Hindi-English (Hi-En) parallel corpus obtained from HindEnCorp0.5 \citep{11858/00-097C-0000-0023-625F-0}. Also, we created smaller training sets from the complete IWSLT14 training set. We randomly sampled 10K, 20K, 30K, 40K, 80K, 120K and 160K sentence pairs from the full training data. The validation and test datasets were the same across all training sets.
{We also evaluated the performance of our method on extremely low resource language pairs. We used Belarusian-English (Be-En) parallel corpus from TED talks \citep{qi2018and} that contains only 4.5K parallel sentences in the training set}.
The HindEnCorp0.5 dataset contains 270K sentence pairs, out of which
we randomly sampled 7K sentence pairs each for validation and test sets, and used the remaining sentences as the training set. 
We used moses toolkit \citep{koehn2007moses} to tokenize German, Belarusian and English sentences, and IndicNLP Library\footnote{\href{https://github.com/anoopkunchukuttan/indic_nlp_library}{IndicNLP Library}} to tokenize Hindi sentences. For open-vocabulary translation, we segmented words using byte-pair encoding (BPE)\footnote{\href{https://github.com/rsennrich/subword-nmt}{https://github.com/rsennrich/subword-nmt}} \citep{sennrich2015neural}. 
{For Be-En parallel corpus, we learned 5K merge operations for both Be and En separately. For other datasets,} we combined the source and target sentences of the training set for learning BPE. We learned 10K merge operations for IWSLT14 dataset, and 20K merge operations for other datasets.

\subsection{Hyperparameter Optimization}

\label{hpmeters}

{The transformer model has a large number of hyperparameters, and hence the total number of combinations of possible values for these hyperparameters is exponential. Therefore, although the language pairs are different from the original pairs used to determine the default values, much of the previous work uses the default hyperparameters (e.g., \citep{gu2018meta, aharoni2019massively}). However, different languages have different characteristics, and using the hyperparameters tuned for one language pair, might not yield the optimal performance for another language pair. Furthermore, the amount of data available for training also affects the choice of hyperparameters. Hence, for each language pair, we perform extensive hyperparameter tuning to get better performance. Since there are exponential number of combinations, grid search is computationally very intensive, and random search is not guaranteed to yield optimal hyperparameters. Hence, we perform hyperparameter search using different values for a subset of hyperparameters. We majorly tune on two types of hyper-parameters - architecture hyper-parameters (e.g., number of attention heads, feed-forward dimension), and regularization hyper-parameters (e.g., dropout, attention dropout, activation dropout, label smoothing). The remaining hyper-parameters such as word embedding size, number of layers, for both encoder and decoder are set to their default values (similar to \citep{vaswani2017attention}), and kept constant throughout the search. We first tune the architecture hyperparameters and keep the regularization hyperparameters constant with their default values. Next, we tune the regularization hyperparameters using the optimal values for architecture hyperparameters. Since we consider only a small subset of hyperparameters, the number of combinations are not exponential, and hence we are able to use grid search to tune the hyperparameters. The optimal hyperparameters chosen are the ones that correspond to the minimum loss on the validation set. Also, we use early stopping (described in Section~\ref{exp_setup_baselines}) to prevent our model from overfitting. 
Although our hyperparameter tuning method does not guarantee a global optimum, we observe a substantial improvement over the default hyperparameters in our experiments (Section~\ref{results}).}
The values of default and optimal hyperparameters obtained using our hyperparameter search, are reported in Table~\ref{hyperparameters}.

\subsection{Experimental Setup and Baselines} \label{exp_setup_baselines}

We consider the Standard Transformer-base model \cite{vaswani2017attention} as a baseline, and for implementation, we used fairseq toolkit \cite{ott2019fairseq}. We also analyzed the effect of applying our proposed approach DHICM to different layers of both encoder and decoder of the transformer model, and observed that
applying the second level attention at the last layer of both encoder and decoder yields the best score.

We refer to the hyperparameters reported in the Standard Transformer-base model \cite{vaswani2017attention} as the Default Hyperparameters, and those {obtained using our hyperparameter search described in Section~\ref{hpmeters}} are referred to as the Optimal Hyperparameters. {We trained all the models on 4 Nvidia GeForce RTX 2080 Ti GPUs. The number of layers of encoder and decoder was set to 6, number of tokens per batch was set to 8000, and the word embedding dimension $d$ was set to 512. We used Adam optimizer ($\epsilon = 10^{-6}, \beta_{1} = 0.9, \beta_{2} = 0.98$)~\citep{kingma2014adam} with a learning rate of $5 \times 10^{-4}$. We used inverse square root learning rate scheduler with 4000 warmup steps, and used beam search with beam size of 5 for generating the sentences. In our proposed approach, we add two additional hyperparameters, that is, $\lambda$ (described in Section~\ref{DHIC}), and a dropout in the second level attention (described in Section~\ref{DHIC}). The optimal values for the dropout added are provided in Table~\ref{hyperparameters}, and we set $\lambda$ as 0.1, for all experiments, corresponding to the minimum loss on the validation set. We save model checkpoints after every epoch and select the best checkpoint based on the lowest validation loss. In order to minimize overfitting, we stop training if the validation loss does not decrease for 10 consecutive epochs.}
For training the models on smaller, randomly sampled training sets from the full IWSLT14 training set (Sec.~\ref{data}), we used the optimal hyper-parameters learned using the full IWSLT14 training set. We used BLEU \citep{papineni2002bleu} as the evaluation metric to compare the performance of our approach with two versions of the baseline model, (i) T-base, which is the Transformer-base model trained using Default hyperparameters, and (ii) T-optimal, which is the Transformer-base model trained using Optimal hyperparameters (Sec.~\ref{hpmeters}). {Please note that, for all our experiments, the hyperparameters for T-optimal and DHICM are same.}

\begin{table}[t]
\centering
{
\begin{tabular}{ccccc}
	\toprule
	Dataset & T-base & T-optimal & DHICM \\
	\midrule
    WMT17-CS & 21.33 & 24.56 & \textbf{25.68}\\
    HindEnCorp & 17.3 & 22.96 & \textbf{26.41} \\
    \specialcell{TED talks \\ (Be-En)} & 4.09 & 5.49 & \textbf{6.29} \\
	\bottomrule
\end{tabular}
}
\caption{BLEU Score of different models on WMT17-CS, HindEnCorp, and Be-En parallel-corpora (trained using full training set). Note that Be-En is an extremely low resource language pair.}
\label{WMT17_HindEnCorp_results}
\end{table}


\section{Results} \label{results}

Table~\ref{WMT17_HindEnCorp_results} shows the performance of different methods. 
We observe that T-optimal outperforms T-base, and this demonstrates that the optimal hyperparameters found in our extensive hyperparameter search yield higher performance compared to the default hyperparameters in \cite{vaswani2017attention}. 
Also, DHICM achieves a higher BLEU score, and outperforms T-optimal on HindEnCorp and WMT17-CS datasets by 3.45 and 1.12 BLEU points, respectively.  
We also performed experiment on the extremely low resource language pair Be-En, and observed that T-base achieved 4.09 BLEU score, and T-optimal achieved 5.49 BLEU score. Thus, T-optimal outperformed T-base by 1.4 BLEU points. Moreover, DHICM achieved 6.29 BLEU score, thus outperforming T-optimal by 0.8 BLEU points. We also compared the performance of our method with the
multi-hop multi-head attention model \cite{iida2019attention} on WMT17-CS De-En dataset. We observed that DHICM outperforms \cite{iida2019attention} by 1.77 BLEU points.

\begin{table}[t]
\centering
{
\begin{tabular}{cccc}
	\toprule
	\specialcell{Train Set Size} & \specialcell{T-base} & \specialcell{T-optimal} & \specialcell{DHICM} \\
	\midrule
    10K & 9.23 & 4.39 & \textbf{14.03} \\
	20K & 13.44 & 7.62 & \textbf{22.00} \\
	30K & 16.43 & 23.06 & \textbf{26.37} \\
	40K & 19.31 & 27.79 & \textbf{28.32} \\
	80K & 27.58 & 32.73 & \textbf{32.93} \\
	120K & 30.93 & 34.60 & \textbf{ 34.7} \\
	160K & 32.72 & 35.85 & \textbf{35.92} \\
	\bottomrule
\end{tabular}
}
\caption{BLEU score averaged over 3 randomly sampled training sets from full IWSLT14 training set}
\label{IWSLT14_results}
\end{table}

\begin{figure*}[!h] 
\centering
\begin{subfigure}[!h]{0.49\textwidth}
    \centering
    \includegraphics[width=\textwidth]{baseline.pdf}
    \caption{Traditional Transformer-Base model}
    \label{baseline}
\end{subfigure}
\begin{subfigure}[!h]{0.49\textwidth}
    \centering
    \includegraphics[width=\textwidth]{aoa.pdf}
    \caption{DHICM}
    \label{aoa}
\end{subfigure}
\caption{Encoder-Decoder Attention distribution (from model trained using 20K sentence pairs of IWSLT14 training set)}
\label{attndist}
\end{figure*}

\begin{figure*}[!h] 
\centering
\begin{subfigure}[!h]{0.49\textwidth}
    \centering
    \includegraphics[width=\textwidth]{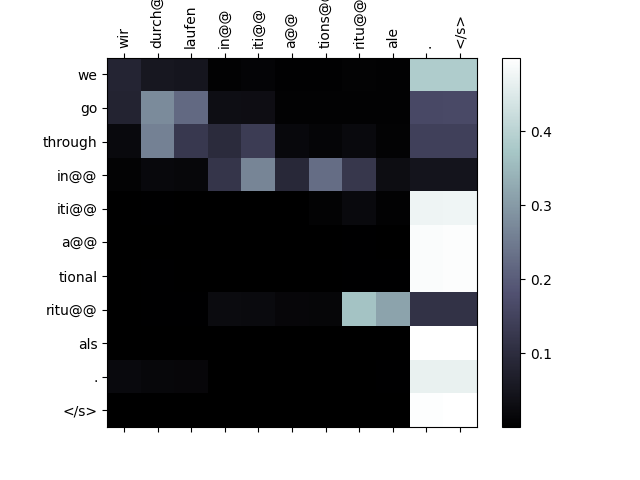}
    \caption{Traditional Transformer-Base model}
    \label{baseline3}
\end{subfigure}
\begin{subfigure}[!h]{0.49\textwidth}
    \centering
    \includegraphics[width=\textwidth]{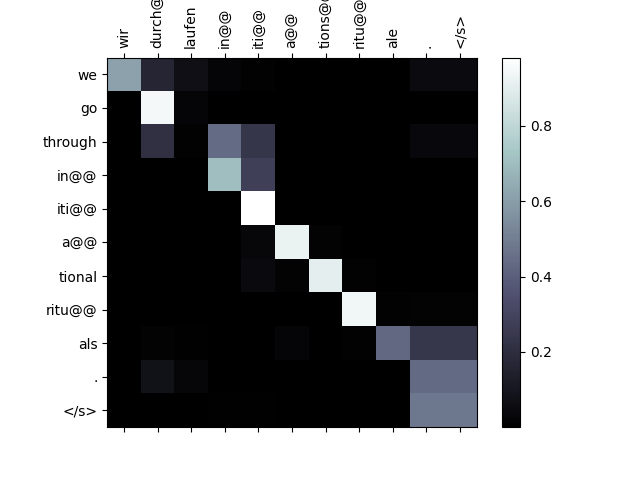}
    \caption{DHICM}
    \label{aoa3}
\end{subfigure}
\caption{Encoder-Decoder Attention distribution (from model trained using 160K sentence pairs of IWSLT14 training set)}
\label{attndist_high}
\end{figure*}

Table~\ref{IWSLT14_results} shows the BLEU score achieved by the models trained with smaller training sets that are randomly sampled from full IWSLT 2014 training set. We observe that the performance of all methods increases with an increase in the training set size, and DHICM achieves a much higher performance compared to T-base for all training set sizes.
The performance of T-optimal and DHICM is similar for larger datasets, however, for low-resource datasets, our approach outperforms T-optimal by a large margin.

Since the hyperaparameters for both T-optimal and DHICM are same, we can see that the gain in the performance of our method is due to the proposed second layer attention over the multi-head attention. In addition, our proposed loss function (Section~\ref{Loss}) 
prevents the model from assigning the same importance to all heads. Thus, we are able to filter more important heads.


\section{Analysis}

Our proposed approach DHICM outperforms T-base and T-optimal by a large margin in the low resource conditions. 
We further analyzed the performance of the baseline model and DHICM, and observed that DHICM learns better word alignment especially, in low resource conditions.
One of the reasons for learning better alignment can be that for each word, all heads are not equally important.
The second level attention that we designed in our model allows the tokens to pay more attention to the heads that capture more relevant information for translation. Since the heads that are more relevant 
receive more attention, the parts of the input to which these heads attend, in turn receive more attention, and thus, the alignment becomes better. For example, providing more attention to the heads that capture the syntactic or semantic information, and relatively less attention to the heads that capture positional information.
This justifies our hypothesis mentioned in Section~\ref{DHIC}.

We also verified this using the encoder-decoder attention distribution of the models shown in Figure~\ref{attndist} (low resource conditions) and Figure~\ref{attndist_high} (high resource conditions). The decoder of the transformer model uses the outputs of the encoder to generate the tokens in the target language. Each generated token pays some attention to each token in the source language. The attention distribution matrix shows the attention paid by the generated tokens in the target sentence (rows) to the tokens in the source sentence (columns). 
In Figure~\ref{baseline} and Figure~\ref{baseline3}, we can see that most of the tokens on the source side get similar attention for the baseline approach. Moreover, the highest attention a source token receives is approximately 0.12 and 0.5 in Figure~\ref{baseline} and Figure~\ref{baseline3}, respectively. This implies that the most important source token for translation does not receive enough attention, resulting in a poor word alignment.
On the contrary, for DHICM (Figure~\ref{aoa} and Figure~\ref{aoa3}), we observe a large variance in the distribution of the attention paid by a target token to the source tokens. Thus, more appropriate source tokens receive higher attention scores $(\sim 0.8)$ in DHICM, leading to a better word alignment, as shown in both Figure~\ref{aoa} and Figure~\ref{aoa3}. Also when 160K training sentences are used for IWSLT14, although the performance of the baseline and DHICM is similar, DHICM learns better word alignments compared to the baseline (shown in Figure~\ref{attndist_high}), as DHICM helps the model to pay more attention to more relevant source tokens. Moreover, DHICM allows the model to pay higher attention ($\sim 0.8$) to the appropriate source words compared to the baseline model where highest attention received by a source token is $\sim 0.5$. This shows that for both low resource and high resource conditions, DHICM helps the model to pay higher attention to the more relevant source tokens.

\begin{figure}[!h] 
\hspace{-10mm}
\centering
    \includegraphics[width=0.4\textwidth]{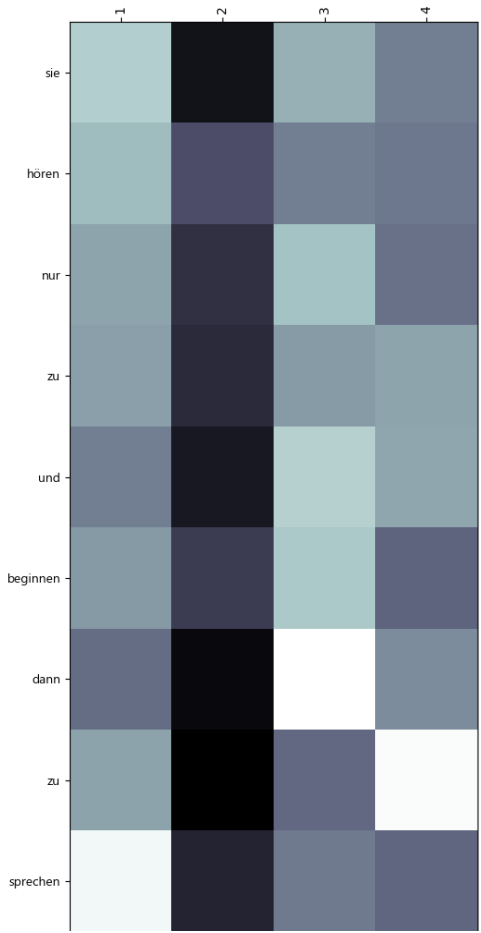}
\caption{Attention paid by each source token to all the heads (from model trained using 160K sentence pairs of IWSLT2014 training set. Brighter/Lighter color corresponds to higher attention and Darker color corresponds to lower attention.}
\label{headattn}
\end{figure}

We also analysed the additional attention layer introduced in DHICM. We compute the attention paid by each token to each head. Using the second level attention, we compute the attention paid by a particular token to all the heads and plot the attention values to create an attention distribution matrix. Figure~\ref{headattn} shows the attention distribution for the second level attention added on top of the multi-head self attention in the last layer of the encoder. The attention distribution matrix shows the attention paid by each source token (rows) to all the 4 heads (columns). The distribution shows that each token pays different amount of attention to each head, and this justifies our hypothesis that all heads are not equally important. Also, different tokens pay different amount of attention to a particular head, which also supports our hypothesis that the importance of a head is dynamic in nature, i.e., it varies as the input token changes. The attention distribution matrix also shows that the additional loss term indeed allows the model to compute different importance scores for different heads. In Figure~\ref{headattn}, we can see that the second head gets the least attention from all the tokens. This shows that our proposed method identifies the least important heads, and thus, by incorporating DHICM, an appropriate pruning strategy can be developed to prune the least important heads.

\section{Related Work}

Some recent work has shown that most of the heads in a multi-head attention model
become redundant during test time 
\citep{michel2019sixteen}. 
\citep{voita2018context, voita2019analyzing} 
analyzed the heads in a multi-head attention model, 
based on some importance score that is calculated after the model is fully trained.
In contrast, in this work, we propose to calculate the importance scores dynamically while training.
%

A recent work \cite{iida2019attention} proposed 
to apply attention on top of the output of multi-head attention. However, they apply an additional attention layer only on the encoder, whereas, in our proposed method, we apply the second level attention on both encoder and decoder, that helps the generated target words to pay significant attention to appropriate source words, which in turn enhances the encoder-decoder attention distribution as shown in Figure~\ref{aoa}. Moreover, their proposed approach might learn equal attention weights for the additional attention layer, which would make all the heads equally important. In such a case, their approach would perform similar to transformer base model, even after adding more number of parameters compared to the standard transformer. To address this, we add an extra loss term in our method, to penalize for learning similar weights for the second level attention. This helps our method to compute different importance scores for different heads. 
Furthermore, during the calculation of the final attention, they transform the output of each head using a different transformation matrix for each head, while our proposed approach DHICM uses a single transformation matrix for the outputs of all heads. Thus, DHICM learns much fewer number of parameters in addition to achieving greater performance 
(the number of additional parameters learned in their approach is 550K, whereas DHICM learns 500K additional parameters). 


\section{Conclusion and Future Work}

In this work, we proposed an effective Dynamic Head Importance Computation Mechanism (DHICM) to dynamically calculate the importance of different heads during training. Our idea is to calculate the importance with an additional attention layer along with the standard multi-head attention. We also proposed a loss function to prevent our method from computing equal importance for all heads, which together with the second-level attention
facilitates to
dynamically identify heads that are most important to the input word. Thus, {the target words generated pay significantly high attention to the more appropriate/relevant source words.}
%
We also performed extensive hyperparameter tuning on a subset of hyperparameters, 
and observed that the optimal hyper-parameters obtained from our search yield a much higher BLEU score compared to the default hyper-parameters. 
Experiments on multiple translation tasks show that DHICM outperforms the standard transformer model 
by a large margin, especially in low resource settings.
In the future, %
we will use the importance scores of the heads computed using DHICM and implement a strategy for pruning the less important heads. We would also like to explore further in the direction of reducing redundancy in multi-head attention.

\bibliographystyle{acl_natbib}
\bibliography{ranlp2021}


\end{document}